\documentclass[sigconf,nonacm]{acmart}
\usepackage{multirow}
\usepackage{tabularx}
\usepackage{algorithm}
\usepackage{algpseudocode}
\usepackage{amsmath}
\usepackage{booktabs}
\usepackage{longtable}
\usepackage{float}
\usepackage{microtype}
\usepackage{textgreek}
\usepackage{tikz}
\usepackage{eso-pic}
\usepackage[utf8]{inputenc}
\usepackage[utf8]{inputenc}

\AtBeginDocument{%
  }

\setcopyright{none}
\acmYear{2027}
\acmConference[KDD '27]
  {33rd ACM SIGKDD Conference on Knowledge Discovery and Data Mining}
  {August 1--5, 2027}
  {San Jose, CA, USA}
\acmBooktitle{Proceedings of the 33rd ACM SIGKDD Conference on Knowledge
  Discovery and Data Mining V.1 (KDD '27),
  August 1--5, 2027, San Jose, CA, USA}
\acmISBN{}
\acmDOI{10.1145/3862868.3871087}

\begin{document}


\title{TabDivER: A Diversity-Enhanced Retrieval Framework for Tabular Foundation Models}

\AddToShipoutPictureFG*{%
  \AtPageUpperLeft{%
    \hspace{1.5cm}%
    \raisebox{-1.5cm}{\small\itshape Accepted at KDD 2027}%
  }%
}
\author{
Haotian Gu$^{1}$,
Hao Liu$^{1}$,
Jiaying Ma$^{1}$,
Jingjun Wang$^{1}$,
Hanqing Zhang$^{1}$,
Haolong Xiang$^{2}$,
Wei Wang$^{3,*}$,
Wenwen Gong$^{1,*}$
\\[0.5em]
\small
$^{1}$China Agricultural University, Beijing, China
\\
$^{2}$Nanjing University of Information Science and Technology, Nanjing, China
\\
$^{3}$Nankai University, Tianjin, China
\\[0.2em]
\small $^{*}$Co-corresponding authors
}

\renewcommand{\shortauthors}{Gu et al.}
\begin{abstract}

Tabular data is one of the most ubiquitous sources of information worldwide, spanning a broad spectrum of domains. The inherent heterogeneity makes generalizing across tables a fundamental bottleneck that has significantly hindered the development of Tabular Foundation Models (TFMs). Recently, in-context learning (ICL) has emerged as a promising solution for TFMs, leveraging a small set of contextual samples to facilitate cross-table inference. 
Existing ICL-based TFMs generally follow two distinct implementation paradigms: retrieval-augmented models and non-retrieval models. However, both paradigms suffer from extremely severe inherent limitations: retrieval-augmented models become overly reliant on highly homogeneous contexts, while non-retrieval models process the full context indiscriminately. Consequently, the inference pipeline becomes entangled with either informational redundancy or semantic noise. The overall performance of ICL-based TFMs remains critically constrained by the quality of context.
To address these challenges, we propose TabDivER, a plug-and-play diversity-enhanced retrieval framework. Specifically, it first constructs a geometry-preserving retrieval space and builds an efficient index for fast candidate retrieval. It then reranks the candidates with a tabular-specific diversity-adaptive component, balancing query similarity with contextual diversity, thereby mitigating distributional bias in the context. Extensive experiments on four established TFMs across three benchmarks (CC18 with 72 tasks, CTR23 with 30 tasks, and Reg\_num with 16 tasks) demonstrate that TabDivER consistently improves TFMs' performance on both classification and regression tasks.


\end{abstract}




\maketitle

\section{Introduction}
\label{intro}
Tabular data constitutes the backbone of most real-world applications, from finance, healthcare, and e-commerce, to many others~\cite{borisov2022deep}. 
This ubiquity has driven decades of research into tabular modeling, making it one of the most actively studied problems in machine learning.
Historically, the landscape of tabular modeling has been dominated by Gradient Boosted Decision Trees (GBDTs), with robust algorithms such as XGBoost ~\cite{chen2016xgboost}, LightGBM ~\cite{ke2017lightgbm}, and CatBoost ~\cite{dorogush2018catboost} serving as the gold standard due to their exceptional performance on heterogeneous features. 
Despite their efficacy, these traditional approaches operate under a task-specific paradigm, requiring individual models to be trained and meticulously tuned for every new dataset. This process requires costly model selection and exhaustive hyperparameter optimization per-dataset~\cite{grinsztajn2022tree,akiba2019optuna,feurer2022auto,mcelfresh2023neural}.

Inspired by the success of Large Language Models (LLMs) ~\cite{vaswani2017attention,brown2020language}, a new generation of Transformer-based Tabular Foundation Models (TFMs), such as TabPFN ~\cite{hollmann2022tabpfn} and TabICL~\cite{qu2025tabicl}, has emerged as a promising solution. At the core of these models is In-Context Learning (ICL)~\cite{kapuriya2025exploring,li2023unified}, which has become a pivotal technology for enabling zero-shot and few-shot cross-domain reasoning. By simply conditioning predictions on a small set of examples, ICL allows TFMs to adapt to diverse schemas and tasks without the need for costly per-dataset fine-tuning or hyperparameter optimization ~\cite{dong2024survey,liu2022makes}. 
This paradigm not only facilitates rapid prototyping but also provides a natural mechanism for handling complex distribution shifts at inference time with significantly less computational overhead than traditional deep learning methods.

Despite its potential, the effectiveness of ICL is fundamentally dependent on the quality of the context provided. However, the vast majority of notable Tabular Foundation Models (TFMs), such as TabPFN and TabICL, lack explicit retrieval mechanisms, often relying on fixed or random examples that fail to provide information-rich contexts. Notably, in tabular ICL, simply increasing the number of random context samples not only fails to improve performance, but can significantly degrade it due to the introduction of label noise and out-of-distribution samples. Moreover, even for models that incorporate retrieval mechanisms in ICL, such as TabDPT and LimiX, the example selection process is predominantly bounded by a relevance-only objective. While proximity ensures local relevance, we argue that excessive homogeneity in the retrieved set introduces severe informational redundancy.

In the sparse, high-dimensional manifolds of tabular data, pure similarity retrieval tends to select neighbors that form a tight cluster, causing the context window to collapse into a single local region. This manifold collapse deprives the model of exposure to structurally diverse samples, forcing it to overfit spurious local correlations and systematically undermining generalization, especially for out-of-distribution test instances~\cite{li-etal-2025-enhancing-retrieval,an-etal-2025-rag}. A natural remedy is to introduce diversity, broadening context coverage beyond the immediate neighborhood. However, tabular data poses unique obstacles that render existing diversity methods from NLP directly inapplicable: heterogeneous feature types, the absence of semantic feature order, and the dimensional collapse of learned representations~\cite{Grinsztajn2022,Gorishniy2021RevisitingDL,Huang2020TabTransformerTD}. Consequently, selecting diversified, information-rich context remains a critically under-explored bottleneck in tabular ICL.

In this paper, we propose \textbf{{TabDivER}}, a model-agnostic and plug-and-play \textbf{\underline{Tab}}ular \textbf{\underline{Div}}ersity-\textbf{\underline{E}}nhanced \textbf{\underline{R}}etrieval framework that can be seamlessly integrated with various tabular foundation models~\cite{hollmann2022tabpfn,qu2025tabicl,ma2025tabdpt,zhang2025limix}. The framework first constructs a geometry-preserving retrieval space through an adaptive feature compilation module without relying on any pre-trained encoder, then establishes an efficient retrieval mechanism within this space to obtain preliminary candidates. The retrieved candidates are subsequently reranked by a tabular-specific diversity-adaptive component. This component operates purely on geometric signals, inherently respecting the permutation-invariant nature of tabular data. By dynamically balancing query similarity and contextual diversity through an adaptive trade-off parameter, it mitigates distributional bias within the context.

The main contributions of this work are summarized as follows:

\vspace{-0.6em}

\begin{itemize}

\item To the best of our knowledge, this is the first
work to systematically investigate retrieval diversity
in tabular in-context learning, revealing limitations
of conventional diversity-based retrieval.

\item We propose TabDivER, a training-free, plug-and-play
framework combining a geometry-preserving retrieval space,
query-adaptive relevance--diversity weighting, and
density-aware redundancy control without per-dataset tuning.

\item We evaluate TabDivER on three benchmarks:
CC18, CTR23, and Reg\_num. Results across four
tabular foundation models demonstrate broad performance
improvements and validate the proposed components.

\end{itemize}

\section{Related Work}
\label{related}
\subsection{Tabular Foundation Models}
Catalyzed by the transformative success of Large Language Models (LLMs), Tabular Foundation Models (TFMs) have emerged as a burgeoning paradigm for cross-domain reasoning and few-shot generalization~\cite{van2024tabular}. Initial research prioritized the adaptation of LLMs via table serialization—transforming structured rows into textual sequences to harness the intrinsic in-context learning (ICL) prowess of pre-trained language priors ~\cite{hegselmann2023tabllm}. These methodologies explore sophisticated fine-tuning on large-scale serialized corpora to encapsulate complex numerical dependencies within the LLM's context window ~\cite{fang2024large}. Concurrently, a specialized class of Transformer architectures, architecturally tailored for tabular modalities, has gained significant prominence. A seminal advancement is TabPFN ~\cite{hollmann2022tabpfn}, which approximates Bayesian inference through pre-training on extensive synthetic datasets. This foundational paradigm has been further augmented by frameworks such as TabDPT ~\cite{ma2025tabdpt} and TabForestPFN ~\cite{breejen2024fine}, which integrate real-world data and tree-based synthetic generators to enrich representation diversity. Modern literature also emphasizes the optimization of the TFM pipeline via architectural refinements like linear attention ~\cite{katharopoulos2020transformers} and context-optimization strategies encompassing prompt tuning and selective data retrieval ~\cite{xu2025mmrag,thomas2024retrieval}. State-of-the-art models, including TabICL~\cite{qu2025tabicl} and TabPFN~\cite{hollmann2025accurate}, continue to extend these frontiers, exhibiting superior scalability and robust predictive performance across heterogeneous tabular domains. 

\vspace{-0.3\baselineskip}
\subsection{Retrieval-based Tabular Models}
\label{retrieval}
The performance of in-context learning is fundamentally dependent on the quality of the context provided. To address this challenge, retrieval-augmented mechanisms have become a critical component. These retrieval approaches construct task-specific, information-rich contexts for each prediction, effectively serving as a non-parametric external memory that enables models to ground their reasoning in directly relevant, instance-level evidence without parameter updates. This paradigm has been successfully integrated into leading tabular foundation models, yielding measurable performance improvements~\cite{hollmann2025accurate, spinaci2025contexttab, xu2024mixture,wen2025scalable}. For instance, LimiX~\cite{zhang2025limix} employs a sophisticated attention-driven retrieval process that evaluates both feature importance and sample-to-sample similarity within its own latent space to curate contexts. Similarly, TabDPT~\cite{ma2025tabdpt} implements retrieval in the raw feature space, utilizing an efficient nearest-neighbor search based on the L2 distance~\cite{thomas2024retrieval}. In both cases, providing highly relevant contextual examples has been shown to significantly improve key metrics such as classification accuracy and regression $R^2$ scores. However, this approach introduces a critical limitation: retrieval mechanisms in LimiX and TabDPT tend to select a narrow set of highly similar samples. This homogeneity can lead models to overfit spurious correlations specific to the retrieved cluster, ultimately impairing their ability to generalize to the full heterogeneity of real-world tabular data. Another line of research integrates retrieval with parametric learning in deep tabular models. TabR~\cite{gorishniy2024iclr-tabr} introduces a differentiable nearest-neighbor mechanism that learns retrieval weights end-to-end via gradient descent on each target dataset. ModernNCA ~\cite{ye2025iclr-revisiting} revisits neighborhood component analysis by replacing fixed distance functions with learnable similarity metrics. While these models benefit from retrieval-augmented expressiveness, they fundamentally operate under the training paradigm: each new dataset requires independent parameter optimization and hyperparameter tuning, and they lack zero-shot cross-domain generalization capability.

In summary, existing ICL-based TFMs exhibit two dominant patterns: models without retrieval (TabPFN-v2, TabICL) rely on random or fixed context and suffer from informational redundancy or noise, while retrieval-augmented models (TabDPT, LimiX) are bounded by similarity-only objectives that induce context homogeneity and impair generalization.
\begin{figure*}[htbp]
  \centering
  \includegraphics[width=\linewidth, keepaspectratio]{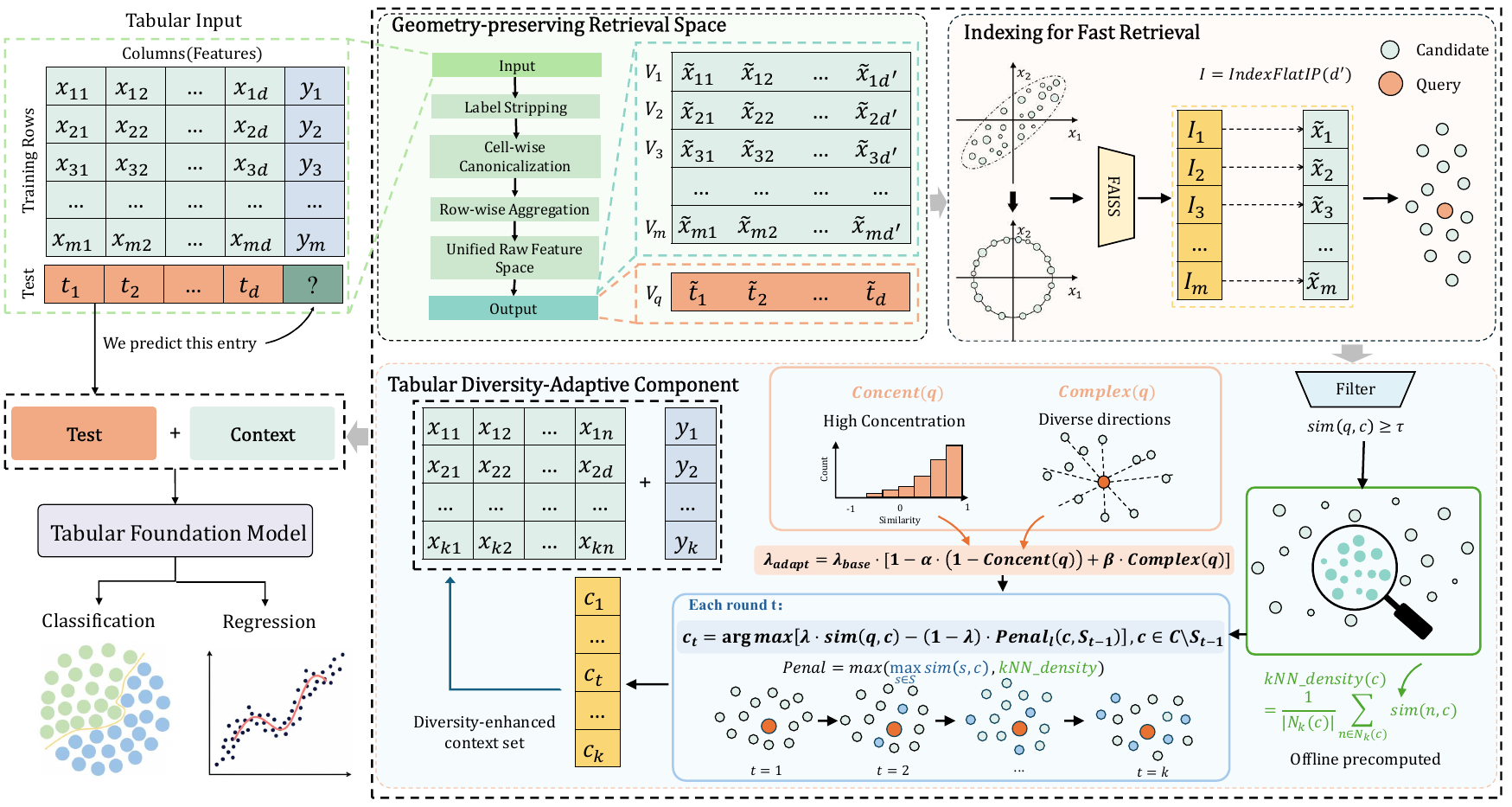}
  \caption{
\textbf{Overview of TabDivER.} The framework consists of three stages. (1) Geometry-Preserving Retrieval Space: heterogeneous tabular features are unified into a geometry-preserving raw feature space via an adaptive feature compilation module. (2) Efficient Candidate Retrieval: features are L$^2$-normalized for cosine similarity search, and a Faiss index retrieves the top-$M$ candidates for each query (orange) to form the initial pool (light green). (3) Tabular Diversity-Adaptive Component: the candidate pool is filtered and reranked by the diversity component with three mechanisms: (i) a local density penalty suppressing over-dense regions, (ii) a query-adaptive \(\lambda\) driven by dual geometric signals, and (iii) a dual-penalty mechanism for set-level and manifold-level redundancy. The final output balances relevance (orange) with diversity (blue) as the TFM context.}
  \label{fig:main1}
\end{figure*}

\section{Methodology}
This section details the three components of TabDivER. We first construct a geometry-preserving retrieval space, then build an efficient index for fast candidate retrieval, and finally introduce a tabular-specific diversity-adaptive component that reranks candidates to balance relevance and diversity.

\subsection{Geometry-preserving Retrieval Space}\label{sec:raw}

The core objective of TabDivER is to explicitly balance
relevance and diversity in tabular in-context learning.
Achieving this goal depends critically on the geometric
integrity of the representation space used for retrieval.
For effective diversity modeling, we consider three
desirable geometric properties: (1) non-degenerate
distance variance, (2) expansible local neighborhood
structure, and (3) preserved effective degrees of freedom.
Formally, consider a $d$-dimensional representation space
with a distance metric $d(\cdot,\cdot)$. For a given query
$q$ and candidate set $C$, if the following holds:

\begin{equation}
    \label{eq:2}
    \operatorname{Var}\bigl[
    d(x_i, x_j) \mid x_i, x_j \in C,\, i \ne j
    \bigr] \approx 0,
\end{equation}

the candidate samples become difficult to distinguish
geometrically, weakening the effectiveness of
distance-based diversity constraints.
Our empirical analysis shows that the unified raw feature
space exhibits lower distance concentration and greater
local neighborhood spread than the encoder-induced space,
providing more distinguishable geometric relationships
for diversity-aware retrieval
(Appendix~\ref{append:raw space}).

Furthermore, diversity-aware retrieval benefits from
expansible local neighborhoods, where successive nearest
neighbors introduce meaningful geometric variation
(Appendix~\ref{append:analysis}).

To maintain this geometric fidelity while handling
heterogeneous data, we propose the following adaptive
unification procedure.

\subsubsection{\textbf{Definition and Label Stripping}} We consider a tabular dataset $D = \{(x_i, y_i)\}_{i=1}^N$ where $x_i$ represents the input features of the $\textit{i}$-th sample, and $y_i$ denotes its corresponding target label. In TabDivER, context retrieval is solely dependent on the geometric relationships between the input features. Therefore, during the construction of the unified feature space, the label $y_i$ is explicitly stripped away to prevent any leakage of label information into the retrieval space. This ensures that the retrieval process remains fully decoupled from downstream prediction tasks. Each sample $x_i$ is composed of $d$ tabular cells $x_i = \bigl(x_{i1}, x_{i2}, \ldots, x_{id}\bigr)$ where the features may belong to various data types, including numerical, categorical, and boolean attributes. TabDivER decomposes the tabular representation problem into two stages: cell-wise value normalization and row-wise vector aggregation. 

\subsubsection{\textbf{Cell-wise Canonicalization}} For numerical features $x_i(k) \in \mathbb{R}$, TabDivER applies feature-wise independent statistical normalization. Missing values are first imputed using the feature mean.
Then we perform z-score standardization for each value: ${x}_i(k)$ is transformed as $\tilde{x}_i(k) = (x_i(k) - \mu_k)/\sigma_k$. 
For categorical and boolean features, we employ a deterministic ordinal mapping function $\tilde{x}_i(k) = \phi_{k}(x_i(k))$, to convert discrete values into integer encodings, with explicit tokens assigned to missing values. 
This avoids dimensional explosion while preserving the distinguishability of categories.
Finally, we systematically remove invalid features, including those with zero-variance, complete missingness, or extremely high cardinality, ensuring that every dimension in the unified space contributes meaningfully to sample differentiation. 

\subsubsection{\textbf{Row-wise Aggregation and Unified Raw Feature Space}}
After cell-wise normalization, each row sample is represented as a real-valued vector
$\tilde{x}_i = (\tilde{x}_{i1}, \tilde{x}_{i2}, \dots, \tilde{x}_{id'}) \in \mathbb{R}^{d'}$, where $d'$ represents the total number of valid features after filtering. This vector is directly determined by the tabular statistical properties and does not rely on any learnable parameters or deep encoders. Hence, it forms a domain-adaptive but non-model-driven raw feature space. It is important to emphasize that this space treats "rows" as the fundamental geometric units. 

To mitigate the dominance of heavy-tailed distributions and extreme outliers in distance calculations, TabDivER introduces a feature-wise $\mu-clipping$ operation in the unified raw space:  
\begin{equation}
    \tilde{x}_i(k) \leftarrow \text{clip}(\tilde{x}_i(k), \mu_k \pm \alpha \sigma_k), 
    \label{eq:6}
\end{equation}which bounds the influence of extreme values in any single dimension. After this process, the unified raw space can be formalized as a metric space $(\mathbb{R}^{d'},d)$ naturally supporting Euclidean distance or cosine similarity, and can be directly used to construct efficient nearest-neighbor indices. 
This design ensures that the unified raw feature space retains the necessary geometric properties for effective diversity-aware retrieval, facilitating robust context selection while maintaining model agnosticism. 

\subsection{Indexing for Fast Retrieval}\label{sec:index}
The second module addresses the computational challenge of efficient nearest-neighbor search within the unified raw feature space constructed in Section~\ref{sec:raw}. It transforms the pre-processed feature vectors into a high-performance, memory-mappable search structure, enabling real-time, diversity-aware retrieval.

The indexing process begins with the canonicalized feature matrix $\mathbf{X} \in \mathbb{R}^{N \times d'}$ from the adaptive unification stage. To enable cosine similarity search, the raw feature vectors are projected onto a unit hypersphere via $L^2$ normalization. This pre-normalization allows direct use of Faiss's IndexFlatIP index, which under $L^2$-normalized inputs becomes functionally identical to cosine search. Normalization is performed via batched, hardware-optimized routines in a single pass, producing $\hat{\mathbf{X}}$.

We employ exact nearest-neighbor search rather than approximate alternatives, prioritizing retrieval accuracy and geometric fidelity. Approximate methods such as hashing or quantization may distort the local neighborhood structures that are crucial for diversity modeling. The index is constructed as $\mathcal{I} = \text{IndexFlatIP}(d')$, where $d'$ is the dimensionality of the normalized vectors, and the entire normalized dataset $\hat{\mathbf{X}}$ is added to the index. The index is serialized with memory-mapping (mmap) support, allowing the operating system to map portions of the index file into virtual memory without loading the entire structure into RAM. For each query, the interface returns the top-$M$ indices and their cosine similarity scores, forming the candidate pool for subsequent diversity-aware reranking in Section~\ref{sec:diversity}.

\subsection{Tabular Diversity-Adaptive Component}\label{sec:diversity}
Building upon the geometry-preserving retrieval space in Section~\ref{sec:raw} and the
efficient retrieval index in Section~\ref{sec:index}, this section
introduces our tabular-specific diversity-adaptive component. We design a component tailored to the geometric structure of tabular data. This mechanism explicitly
balances relevance and diversity in tabular context selection. Given a query sample $\mathbf{\textit{q}} \in \mathbb{R}^{d'}$ and a candidate set $\mathcal{C} = { \mathbf{c}_1, \mathbf{c}_2, \dots, \mathbf{c}_N }$, our objective is to select $k$ samples $S \subseteq \mathcal{C}$ that simultaneously ensure relevance to the query while maximizing diversity among the selected samples. We formalize this objective as the following optimization problem:
\begin{equation}
\max_{S \subseteq C, |S| = k} \big\{ \lambda R(S, \mathbf{\textit{q}}) - (1 - \lambda) D(S) \big\}, 
\label{eq:11}
\end{equation}
where $R(S, \mathbf{\textit{q}})$ is the average relevance score and $D(S)$ measures the internal similarity (negative diversity), defined as:
\begin{equation}
R(S, \mathbf{\textit{q}}) = \frac{1}{|S|} \sum_{\mathbf{c} \in S} \text{sim}(\mathbf{\textit{q}}, \mathbf{c}), 
\label{eq:12}
\end{equation}
\begin{equation}
D(S) = \frac{1}{|S|(|S| - 1)} \sum_{\substack{\mathbf{c}_i, \mathbf{c}_j \in S \\ i \neq j}} \text{sim}(\mathbf{c}_i, \mathbf{c}_j), \label{eq:13}
\end{equation}
where $\text{sim}(\cdot, \cdot)$ denotes the cosine similarity function, and $\lambda \in [0, 1]$ is a balancing parameter. When $\lambda = 1$, the algorithm degenerates to traditional similarity-ranked retrieval; when $\lambda = 0$, it pursues maximal diversity while disregarding relevance. 

Standard Maximum Marginal Relevance ~\cite{carbonell1998use} reduces set-level redundancy by penalizing similarity to already-selected items. However, its fixed $\lambda$ cannot achieve a principled relevance-diversity balance: tabular queries face widely varying data distributions, each requiring a different optimal tradeoff. Moreover, set-level redundancy reduction alone is insufficient for tabular data---the manifold collapse documented in Section~\ref{sec:raw} concentrates samples in over-dense clusters that a set-level penalty cannot discriminate. TabDivER addresses these limitations through a fundamentally different penalty function: we introduce a local density penalty via self-kNN (Eq.~\ref{eq:17}) that explicitly suppresses samples from over-dense manifold regions, together with a query-adaptive $\lambda$ (Eq.~\ref{eq:10}) driven purely by geometric signals, which dynamically adjusts the relevance-diversity balance based on each query's candidate distribution and eliminates per-dataset tuning.

Since the combinatorial optimization in Equation~\ref{eq:11} is NP-hard, we employ a greedy algorithm to approximate the solution. The core idea of the algorithm is to iteratively construct the result set by selecting, at each step, the sample that maximizes the marginal gain. We design a three-stage retrieval strategy:

\textbf{Stage 1}: Similarity-based pre-screening. We first retrieve the $M$ candidate samples most similar to the query, where $M$ is chosen based on the available retrieval samples and target context size: $\mathcal{C}_{pre}={c_{i_1}, c_{i_2}, . . . , c_{i_M}}$. Here, $\text{sim}(\mathbf{\textit{q}}, \mathbf{c}_{i_1}) \geq \text{sim}(\mathbf{\textit{q}}, \mathbf{c}_{i_2}) \geq \dots \geq \text{sim}(\mathbf{\textit{q}}, \mathbf{c}_{i_M})$. We set the candidate-pool size according to the number of available retrieval samples $N$ and the target context size $k$, using $M=\min\{N,15k\}$. This retains the original $15k$ upper bound while avoiding a fixed candidate count that exceeds the available samples on smaller datasets. 

\textbf{Stage 2}: Low-quality sample filtering. To further enhance retrieval quality, we introduce a similarity threshold $\tau$ to filter out low-quality candidates: $\mathcal{C}_{filtered}=\left\{c\in \mathcal{C}_{pre}|sim(q, c) \ge \tau\right\}$. In our experiments, we set $\tau = 0. 3$, which effectively filters out noisy samples while preserving sufficient diversity. In the following, we denote the filtered candidate set \(\mathcal{C}_{\text{filtered}}\) simply as \(C\) for notational convenience. 

\textbf{Stage 3}: Iterative candidate expansion. In subsequent iterations, the result set $S$ is expanded greedily. Given the set of previously selected samples $S_{t-1} = \{\mathbf{c}_1, \dots, \allowbreak \mathbf{c}_{t-1}\}$. To resolve the "manifold collapse" problem inherent in tabular data, we design a geometric-signal-based penalty by introducing a local neighborhood constraint that captures both set-level redundancy and local neighborhood density. For each candidate $c$ from the remaining pool $\mathcal{C} \setminus S_{t-1}$, we define its local diversity penalty as:
\begin{flalign}
    &\text{Penal}_{l}(c, S)=\max\biggl(\max\limits_{s\in S}\text{sim}(s, c), \frac{1}{|N_k(c)|}\sum_{n\in N_k(c)}\text{sim}(n, c)\biggr), 
    \label{eq:17}
\end{flalign}
where the components are defined as follows: The $N_k(c)$ for each candidate is obtained via a self-kNN search within the unified raw feature space. 
Utilizing the high-performance Faiss index constructed in Section 3.2, we pre-compute the average similarity of each sample to its $k$ nearest neighbors, \textbf{\(n \in N_k(c)\)} represents a neighbor of \(c\) in its local neighborhood. 
The second term computes the average similarity between \(c\) and its neighbors, serving as a local density estimator. 
By employing the \(\max(\cdot)\) operator, TabDivER ensures a candidate is penalized if it is either redundant to the current set \(S\) or resides in a hyper‑dense region of the data manifold. 
Accordingly, the algorithm selects the \(t\)-th sample \(c_t\) from the remaining candidates as:
\begin{flalign}
    c_t=\arg \max\limits_{c\in \mathcal{C}\setminus\mathcal{S}_{t-1}}[\lambda \cdot \text{sim}(q, c)-(1-\lambda)\cdot Penal_{l}(c, S_{t-1})]. 
    \label{eq:15}
\end{flalign}
This selection criterion intuitively captures the essence of diversity component: the first term $\lambda \cdot \text{sim}(q, c)$ encourages the selection of samples with high query relevance, while the second term $(1 - \lambda) \cdot \text{Penal}_l(c, S_{t-1})$ penalizes candidates based on their contextual redundancy and geometric density. This dual-penalty strategy allows TabDivER to curate a context window that is both locally faithful to the query and globally representative of the underlying data distribution. 

Rather than employing a fixed value, we propose a query-adaptive tuning method for the trade-off parameter $\lambda$. This mechanism modulates the balance between relevance and diversity based on two geometric signals: similarity distribution concentration and local structural complexity. The unnormalized concentration is defined as $\mathrm{Concent}_{\mathrm{raw}}(q)=\mu_{\mathrm{sim}}/\sigma_{\mathrm{sim}}$, where $\mu_{\mathrm{sim}}$ and $\sigma_{\mathrm{sim}}$ are the mean and standard deviation of the similarities among the top-$L$ candidates. Structural complexity is measured by their mean Euclidean distance from the centroid $\bar{\mathbf c}=\frac{1}{L}\sum_{i=1}^{L}\mathbf c_i$:
\begin{equation}
    \mathrm{Complex}_{\mathrm{raw}}(q)=\frac{1}{L}\sum_{i=1}^{L}\|\mathbf c_i-\bar{\mathbf c}\|_2.
    \label{eq:9}
\end{equation}
The implementation applies bounded monotonic transformations involving $\tanh(\cdot)$ and the sigmoid function $\sigma(x)=1/(1+e^{-x})$ to the two geometric signals before computing the adaptive weight. We denote the resulting normalized quantities by $\widehat{\mathrm{Concent}}(q)$ and $\widehat{\mathrm{Complex}}(q)$. These transformations reduce sensitivity to different scales and extreme values while preserving the ordering of the corresponding signals. The query-specific weight is then computed as
\begin{equation}
    \lambda_{\mathrm{adapt}}
    = \lambda_{\mathrm{base}}
    \left[
    1-\alpha\left(1-\widehat{\mathrm{Concent}}(q)\right)
    +\beta\widehat{\mathrm{Complex}}(q)
    \right],
    \label{eq:10}
\end{equation}
where $\lambda_{\mathrm{base}}$ is the baseline relevance weight and $\alpha,\beta$ are fixed nonnegative coefficients (specified in Algorithm~\ref{alg:diversity_component}). A lower concentration reduces $\lambda_{\mathrm{adapt}}$, placing greater emphasis on diversity, whereas higher local complexity increases the relevance weight. The goal is to adapt the trade-off across queries without selecting a separate $\lambda$ on each dataset.

\section{Experiments}

\subsection{Baselines}
\label{baseline}
To comprehensively evaluate TabDivER, we compare it with four state-of-the-art ICL-based TFMs. TabPFN-v2~\cite{hollmann2025accurate} performs Bayesian inference without explicit retrieval. TabICL~\cite{qu2025tabicl} uses a Transformer with an extended context window but lacks retrieval, relying on random context selection. It supports classification only and cannot be applied to regression tasks. LimiX~\cite{zhang2025limix} employs attention-based retrieval in its latent space. TabDPT~\cite{ma2025tabdpt} retrieves context via Euclidean distance in a raw feature space. We further benchmark against two tree-based models (XGBoost~\cite{chen2016xgboost}, LightGBM~\cite{ke2017lightgbm}) and two retrieval-based deep tabular models (TabR~\cite{gorishniy2024iclr-tabr}, ModernNCA~\cite{ye2025iclr-revisiting}). The former represent the classical GBDT paradigm, while the latter represent the parametric learning paradigm—all require per-dataset training or fine-tuning.
\subsection{Datasets and Evaluation Metrics}
\label{dataset}
We evaluate TabDivER across three established benchmarks spanning diverse task modalities. The OpenML-CC18 ~\cite{bischl2017openml} comprises 72 classification tasks with varying sample sizes, feature dimensionalities, and class distributions. The Reg\_num from Grinsztajn et al.~\cite{grinsztajn2022tree} provides 16 regression tasks covering heterogeneous real-world domains. The CTR23 ~\cite{bischl2017openml} contributes 30 regression tasks, further testing generalization under extreme sparsity and high-cardinality categorical features. For classification, we report Accuracy and ROC AUC; for regression, we report $R^2$ and RMSE.

\begin{table*}[t]
\caption{Evaluation results across OpenML-CC18, CTR23, and Reg\_num benchmarks.}
\label{tab:main_results}
\centering
\normalsize
\begin{tabular}{@{}lc@{\hskip 3pt}c@{\hskip 3pt}c@{\hskip 3pt}c@{\hskip 3pt}c@{\hskip 3pt}c@{\hskip 3pt}c@{\hskip 3pt}c@{\hskip 3pt}c@{\hskip 3pt}c@{}}
\toprule
\multirow{2}{*}{Method} & \multicolumn{3}{c}{OpenML-CC18} & \multicolumn{3}{c}{CTR23} & \multicolumn{3}{c}{Reg\_num} \\
\cmidrule(lr){2-4}\cmidrule(lr){5-7}\cmidrule(lr){8-10}
 & ACC ($\uparrow$) & AUC ($\uparrow$) & Rank ($\downarrow$) & R$^2$ ($\uparrow$) & RMSE ($\downarrow$) & Rank ($\downarrow$) & R$^2$ ($\uparrow$) & RMSE ($\downarrow$) & Rank ($\downarrow$) \\
\midrule
\multicolumn{10}{@{}l}{\textbf{ICL-based TFMs}} \\
\quad TabPFN-v2 & 0.827 $\pm$ 0.002 & 0.859 $\pm$ 0.005 & 11.45 & 0.708 $\pm$ 0.003 & 0.512 $\pm$ 0.004 & 7.21 & 0.736 $\pm$ 0.002 & 0.505 $\pm$ 0.003 & 8.17 \\
\quad +TabDivER & \textbf{0.853 $\pm$ 0.004} & \textbf{0.874 $\pm$ 0.007} & 8.69 & \textbf{0.730 $\pm$ 0.005} & \textbf{0.501 $\pm$ 0.003} & 3.17 & \textbf{0.751 $\pm$ 0.003} & \textbf{0.494 $\pm$ 0.005} & 4.05 \\
\quad TabICL & 0.871 $\pm$ 0.003 & 0.926 $\pm$ 0.001 & 5.24 & --- & --- & --- & --- & --- & --- \\
\quad +TabDivER & \textbf{0.888 $\pm$ 0.005} & \textbf{0.930 $\pm$ 0.005} & 2.49 & --- & --- & --- & --- & --- & --- \\
\quad TabDPT & 0.872 $\pm$ 0.007 & 0.922 $\pm$ 0.005 & 4.12 & 0.720 $\pm$ 0.006 & 0.496 $\pm$ 0.008 & 5.10 & 0.749 $\pm$ 0.005 & 0.484 $\pm$ 0.004 & 5.02 \\
\quad +TabDivER & \textbf{0.887 $\pm$ 0.008} & \textbf{0.928 $\pm$ 0.006} & 3.05 & \textbf{0.737 $\pm$ 0.007} & \textbf{0.492 $\pm$ 0.004} & 2.19 & \textbf{0.759 $\pm$ 0.006} & \textbf{0.478 $\pm$ 0.008} & 2.57 \\
\quad LimiX-16M & 0.878 $\pm$ 0.002 & 0.927 $\pm$ 0.003 & 3.67 & 0.729 $\pm$ 0.001 & 0.490 $\pm$ 0.001 & 3.52 & 0.752 $\pm$ 0.003 & 0.481 $\pm$ 0.001 & 4.64 \\
\quad +TabDivER & \textbf{0.888 $\pm$ 0.001} & \textbf{0.939 $\pm$ 0.005} & 1.60 & \textbf{0.739 $\pm$ 0.003} & \textbf{0.488 $\pm$ 0.004} & 2.03 & \textbf{0.761 $\pm$ 0.005} & \textbf{0.476 $\pm$ 0.003} & 1.82 \\
\midrule
\multicolumn{10}{@{}l}{\textbf{Retrieval-based Deep Tabular Models}} \\
\quad TabR & 0.842 $\pm$ 0.009 & 0.869 $\pm$ 0.007 & 10.48 & 0.676 $\pm$ 0.008 & 0.543 $\pm$ 0.010 & 9.14 & 0.704 $\pm$ 0.008 & 0.523 $\pm$ 0.009 & 10.14 \\
\quad ModernNCA & 0.839 $\pm$ 0.005 & 0.872 $\pm$ 0.006 & 11.83 & 0.675 $\pm$ 0.004 & 0.539 $\pm$ 0.007 & 8.69 & 0.711 $\pm$ 0.005 & 0.527 $\pm$ 0.004 & 9.02 \\
\midrule
\multicolumn{10}{@{}l}{\textbf{Tree-based Models}} \\
\quad XGBoost & 0.873 $\pm$ 0.006 & 0.916 $\pm$ 0.008 & 6.07 & 0.712 $\pm$ 0.007 & 0.509 $\pm$ 0.005 & 6.05 & 0.739 $\pm$ 0.007 & 0.499 $\pm$ 0.006 & 6.07 \\
\quad LightGBM & 0.872 $\pm$ 0.004 & 0.913 $\pm$ 0.005 & 6.93 & 0.708 $\pm$ 0.006 & 0.516 $\pm$ 0.004 & 7.38 & 0.737 $\pm$ 0.003 & 0.501 $\pm$ 0.005 & 6.86 \\
\bottomrule
\end{tabular}
\end{table*}

\subsection{Parameter Setting and Complexity Analysis}

TabDivER selects $K \in \{512, 1024, 2048\}$ context examples using a query-adaptive relevance--diversity coefficient $\lambda$ (Eq.~\ref{eq:10}). The same adaptive weighting rule is applied across all datasets without dataset-specific tuning. Detailed parameter settings are provided in Algorithm~\ref{alg:diversity_component} in the Appendix.

The framework uses a Faiss \texttt{IndexFlatIP} index over the unified feature space. Building the index requires $\mathcal{O}(Nd')$ time and storage. For each query, exact inner-product search computes similarities against all $N$ indexed vectors at a cost of $\mathcal{O}(Nd')$, followed by top-$M$ candidate selection. By incrementally maintaining each candidate's maximum similarity to the selected set, the greedy reranking process requires $\mathcal{O}(MK)$ candidate-score updates, rather than $\mathcal{O}(MK^2)$ under naive recomputation. This update count excludes the feature-level cost of similarity evaluations and offline density estimation. Further complexity details are provided in Appendix~\ref{appendix:complexity}.

\section{Experimental Results and Discussions}

\subsection{Main Experimental Results}
\label{main data}
Table~\ref{tab:main_results} reports aggregated results across the three benchmark suites. On classification tasks, TabDivER consistently improves both ACC and AUC over all four ICL-based TFMs. LimiX-16M achieves the highest absolute performance (ACC 0.888, AUC 0.939 with TabDivER), while TabPFN-v2 exhibits the largest relative gain (+2.5\% ACC, +1.4\% AUC), benefiting most from diversity-enhanced retrieval to compensate for its lack of a built-in retrieval mechanism. TabICL and TabDPT also show clear improvements, with TabICL's ACC rising from 0.871 to 0.888. Among non-ICL baselines, XGBoost and LightGBM outperform several TFM baselines but are consistently surpassed by all +TabDivER variants; TabR and ModernNCA trail further behind. On the CTR23 and Reg\_num benchmarks, TabDivER delivers consistent R$^2$ improvements across all applicable models. On CTR23, TabPFN-v2 shows the largest R$^2$ gain (+2.2\%). On Reg\_num, similar trends hold, with notable gains for TabPFN-v2 (+1.5\%). Tree-based models (XGBoost, LightGBM) remain competitive with TFM baselines but are consistently outperformed by TabDivER-enhanced variants. Overall, the experimental results validate the effectiveness of tabular-specific diversity-enhanced retrieval as a general-purpose plug-in for tabular ICL.
Despite the high baseline efficiency of similarity search in these compact spaces, TabDivER still provides consistent, model-dependent improvements (e. g. , ~1\% for TabICL and LimiX), demonstrating the universal value of balancing similarity and diversity. 

\subsection{Ablation Study}
\label{ablation}
To systematically deconstruct the contributions of TabDivER, we conduct two sets of ablation experiments: retrieval space ablation and diversity component ablation. All experiments are performed on CC18 using TabPFN-v2 with $K=2048$. The results are summarized in Table~\ref{tab:ablation}.

\noindent\textbf{Retrieval space ablation.} Compares our geometry-preserving feature space against the learned embedding space from the TabICL encoder. When TabDivER is applied in the embedding space, accuracy drops sharply from 0.853 to 0.658 (a degradation of 0.19), and AUC from 0.874 to 0.726. This confirms that the learned encoder suffers from representation collapse, which renders diversity-aware retrieval ineffective, thereby validating our design choice of operating in the original feature space.

\begin{table}[htbp]
\centering
\caption{Ablation study on CC18.}
\label{tab:ablation_results}
\small
\begin{tabular}{@{}l|c|c|c@{}}
\toprule
\multirow{2}{*}{Metric} & Retrieval Space & \multicolumn{2}{c}{Diversity Component} \\
\cmidrule(lr){2-2}\cmidrule(lr){3-4}
 & Ours over Emb & Ours over Random & Ours over Similarity \\
\midrule
ACC & +0.194 & +0.201 & +0.021 \\
AUC & +0.148 & +0.278 & +0.009 \\
\bottomrule
\end{tabular}
\label{tab:ablation}
\end{table}

\begin{figure}[htbp]
  \centering
  \includegraphics[width=\linewidth]{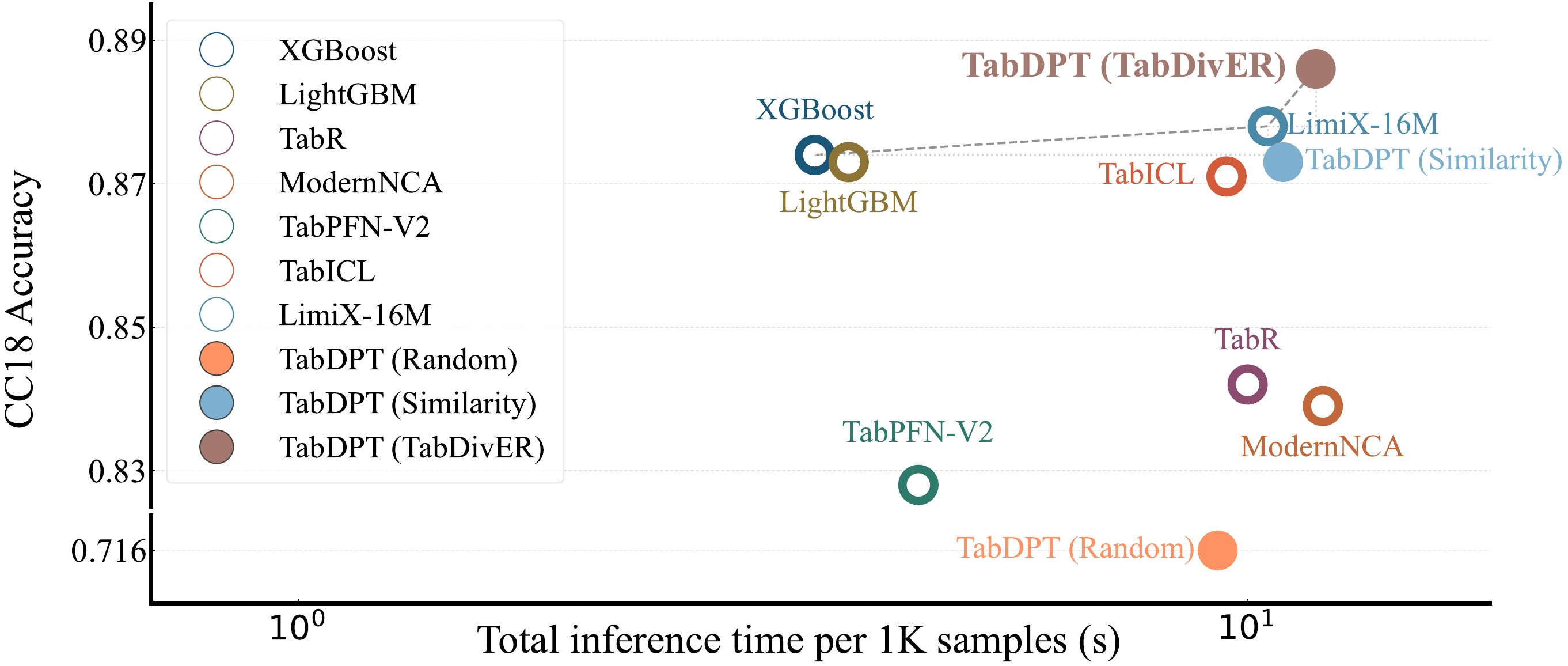}
  \caption{Accuracy vs. total inference time per 1000 samples on CC18 (TabDPT, K=2048). The dashed line indicates the Pareto frontier, where TabDivER achieves the optimal accuracy-latency trade-off. }
  \label{fig:pareto}
\end{figure}

\noindent\textbf{Diversity component ablation.} Evaluates TabDivER against two baselines: random context selection and similarity-only retrieval. Compared to random selection, TabDivER improves ACC by 0.201 and AUC by 0.278, demonstrating that structured retrieval is far superior to undirected sampling. Compared to similarity-only retrieval, TabDivER achieves a consistent gain of +0.021 in ACC and +0.009 in AUC, confirming that beyond relevance-driven selection, diversity provides additional value by reducing redundancy and broadening the coverage of the context window. Figure~\ref{fig:pareto} presents a Pareto-front analysis comparing accuracy against total inference time per 1000 samples. TabDivER achieves the Pareto frontier, reaching the highest accuracy with only a moderate latency increase over similarity-only retrieval.

Standard diversity strategies from NLP and information retrieval cannot be directly applied to tabular data because they rely on semantic embeddings that tabular data lacks. We therefore evaluate all methods within TabDivER's geometry-preserving retrieval pipeline (Stage 1--2), ensuring that every strategy operates over the same retrieval space and candidate pool. This isolates the contribution of the reranking mechanism (Stage 3). Direct application to the original feature space is often ineffective, however, due to the lack of semantic embeddings and representation collapse in learned encoders for tabular data. Unlike general-purpose methods relying on pre-trained embeddings, TabDivER is designed from first principles for tabular geometry via three innovations: a local density penalty, a query-adaptive $\lambda$ from dual geometric signals, and a dual-penalty mechanism for set-level redundancy and manifold-level density.

As shown in Table~\ref{tab:diversity_compare} (TabPFN-v2, $k$=2048), Uniform sampling performs poorly as it disregards both query relevance and sample diversity. K-means centroid selection and Max-Min diversification offer moderate improvements but remain query-agnostic or outlier-sensitive, often selecting noisy samples in high-dimensional tabular spaces. MMR ($\lambda=0.5$) uses a fixed $\lambda$ and
underperforms similarity-only retrieval
(0.816 vs. 0.832 ACC).
TabDivER achieves the highest accuracy (0.853),
significantly outperforming all other methods,
while maintaining competitive latency (14.6 ms/query)
and high throughput (1,980 QPS) through GPU-parallel
batched query processing.
Although DPP~\cite{Kulesza2012DeterminantalPP}
provides a theoretically principled diversity metric,
its cubic complexity $O(M^3)$ renders it
prohibitively expensive for real-time ICL deployment.

\begin{table}[htbp]
\centering
\caption{Comparison of diversity strategies on CC18.}
\small
\begin{tabular}{@{}l|c|c|c|c@{}}
\toprule
Method & ACC & AUC & Latency (ms/query) & QPS \\
\midrule
Uniform  & 0.713 & 0.808 & $\sim$1.1 & $>$10,000 \\
K-means ~\cite{MacQueen1967SomeMF}& 0.747 & 0.820 & $\sim$6.2 & $\sim$3,000 \\
Max-Min ~\cite{Gonzalez1985ClusteringTM} & 0.761 & 0.837 & $\sim$13.3 & $\sim$620 \\
MMR ~\cite{Carbonell1998TheUO} & 0.816 & 0.844 & $\sim$13.9 & $\sim$2,100 \\
\midrule
TabDivER (Ours) & \textbf{0.853} & \textbf{0.874} &$\sim$ 14.6 & $\sim$1,980 \\
\bottomrule
\end{tabular}
\label{tab:diversity_compare}
\end{table}

\noindent\textbf{Adaptive weighting and density ablation.}
To isolate the contributions of the proposed components,
we evaluate four variants using the same retrieval space
and candidate pool (Table~\ref{tab:component_ablation}).
Similarity-only retrieval achieves 0.832 ACC, while vanilla
MMR (fixed $\lambda=0.5$, without adaptive weighting or
density) drops to 0.816, showing that a fixed diversity
trade-off can be detrimental. Enabling query-adaptive
weighting increases ACC to 0.844 (+2.8 percentage points
over vanilla MMR), and adding the local-density penalty
further raises it to 0.853 (+0.9 percentage points).
Thus, the full method improves over similarity-only
retrieval by 2.1 percentage points; the gains stem from
adaptive balancing and density-aware redundancy control
rather than diversity alone.

\begin{table}[htbp]
\centering
\caption{Component-wise ablation on CC18.}
\label{tab:component_ablation}
\small
\begin{tabular*}{0.90\columnwidth}
{@{\extracolsep{\fill}}lccc@{}}
\toprule
Method & $\lambda$ & ACC & AUC \\
\midrule
Similarity-only & $1$ & 0.832 & 0.865 \\
w/o Adaptive \& Density & $0.5$ & 0.816 & 0.844 \\
w/o Density & Adaptive & 0.844 & 0.862 \\
\midrule
Full TabDivER & Adaptive &
\textbf{0.853} & \textbf{0.874} \\
\bottomrule
\end{tabular*}
\end{table}

\subsection{Parameter Sensitivity Analysis}
\label{para}

This section examines how TabDivER's key hyperparameters,
specifically the relevance--diversity trade-off coefficient
$\lambda$ and the context size $K$, affect downstream task
performance. We conduct a grid search over
$\lambda \in [0.1, 0.8]$ and
$K \in \{512, 1024, 2048\}$, averaging results across
four TFM backbones on CC18 (ACC) and CTR23 ($R^2$).

\begin{figure}[htbp]
  \centering
  \includegraphics[width=\linewidth]{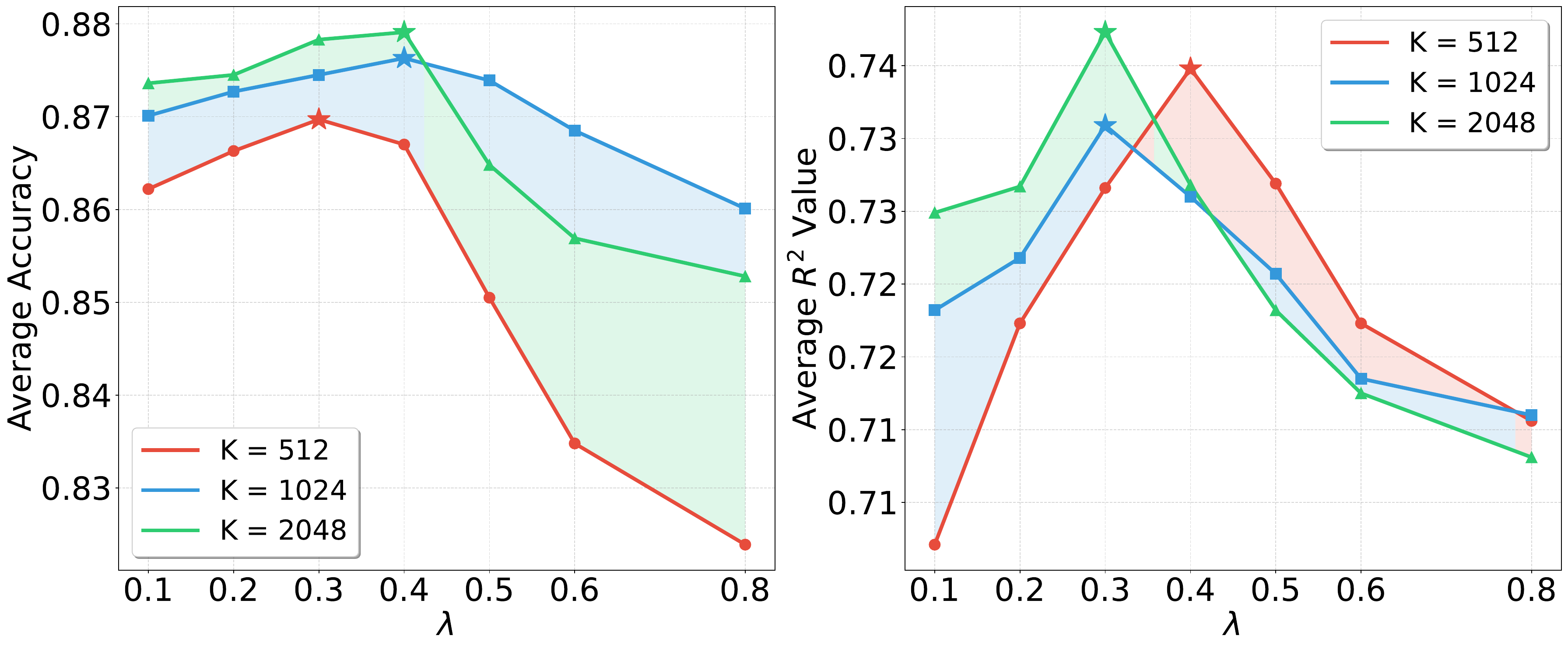}
  \caption{Impact of parameters ($\lambda$ and $K$)
  on performance.}
  \label{fig:parameters}
\end{figure}

\noindent\textbf{Optimal diversity interval.}
As shown in Figure~\ref{fig:parameters}, performance
generally peaks at moderate $\lambda$ values, approximately
between $0.3$ and $0.4$ in the evaluated grid.
The adaptive $\lambda$ mechanism (Eq.~\ref{eq:10})
produces query-specific weights ranging from $0.26$ to
$0.74$ on CC18, rather than relying on a single fixed
value. When $\lambda$ is too low ($\leq 0.2$), excessive
diversity can introduce less relevant samples that dilute
local signals; when $\lambda$ is too high, excessive
emphasis on relevance can produce redundant contexts,
degrading predictive performance. This highlights the
importance of balancing relevance and diversity without
dataset-specific tuning.

\vspace{1em}

\noindent\textbf{Context size scaling.}
Overall, larger context windows tend to improve
performance by providing richer contextual information.
However, we observe a notable inverse scaling phenomenon
under certain higher-relevance settings.
For example, at $\lambda=0.5$, ACC drops from
$0.874$ ($K=1024$) to $0.865$ ($K=2048$).
This suggests that when relevance is strongly emphasized,
enlarging the context window may introduce additional
redundancy and diminishing returns, making a more
compact context more effective.

\subsection{Robustness to Temporal Distribution Shift}
\label{robust}
TabReD~\cite{rubachev2025iclr-tabred} is a benchmark of industrial-scale tabular datasets with temporal splitting, where training precedes test in time, causing natural distribution shift. Under this challenging setting, similarity-based retrieval often selects historically similar but currently outdated samples, limiting its effectiveness. Following prior work, we exclude HomeCredit Default and Maps Routing due to their feature dimensionality exceeding 500, which exceeds the input limits of our baseline TFMs. We evaluate TabDivER on the remaining six TabReD datasets to assess its robustness under temporal shift. Detailed results for these datasets are presented in Table~\ref{tab:tabred}.

\begin{table}[htbp]
\centering
\caption{Results on TabReD under temporal split.}
\label{tabred}
\small
\begin{tabular}{@{}l@{\hskip 10pt}c@{\hskip 10pt}c@{\hskip 10pt}c@{\hskip 10pt}c@{\hskip 10pt}c@{}}
\toprule
Method & ACC & AUC & R$^2$ & RMSE & Rank \\
\midrule
TabPFN-v2 & 0.844 & 0.904 & 0.495 & 0.692 & 8.14 \\
+TabDivER & \textbf{0.850} & \textbf{0.909} & \textbf{0.508} & \textbf{0.685} & 7.43 \\
TabICL & 0.860 & 0.919 & --- & --- & --- \\
+TabDivER & 0.856 & 0.916 & --- & --- & --- \\
TabDPT & 0.846 & 0.907 & 0.509 & 0.681 & 7.86 \\
+TabDivER & \textbf{0.859} & \textbf{0.913} & \textbf{0.517} & \textbf{0.675} & 5.29 \\
LimiX-16M & 0.867 & 0.922 & 0.532 & 0.663 & 2.57 \\
+TabDivER & 0.865 & 0.916 & 0.532 & 0.667 & 3.43 \\
\midrule
XGBoost & 0.849 & 0.909 & 0.494 & 0.684 & 7.02 \\
LightGBM & 0.853 & 0.910 & 0.501 & 0.683 & 6.64 \\
TabR & 0.829 & 0.899 & 0.482 & 0.699 & 10.29 \\
ModernNCA & 0.831 & 0.902 & 0.478 & 0.705 & 10.86 \\
\bottomrule
\end{tabular}
\label{tab:tabred}
\end{table}

Under temporal distribution shift, tabular data exhibits a retrieval drift phenomenon: structurally similar samples can carry diverging labels as feature-label relationships evolve over time. This creates distinct dynamics across model types. For TabPFN-v2, whose small context window limits random sampling coverage, TabDivER provides a net gain by broadening candidate scope. For TabDPT, whose L2 nearest-neighbor retrieval (0.846 ACC) selects temporally stale samples, diversity compensates for this retrieval bias. For TabICL, random sampling over a much larger context window already provides implicit diversity; introducing explicit retrieval disrupts this unbiased coverage, degrading performance. For LimiX, attention-driven retrieval already selects highly relevant context, and additional diversity introduces noise from label-irrelevant regions. This pattern reveals a fundamental tradeoff: diversity-enhanced retrieval is most effective when context selection is simple or the context window is constrained, while models with implicit diversity or strong built-in retrieval can be negatively affected. Understanding this interaction under concept drift is left to future work.

\begin{figure}[htbp]
  \centering
  \includegraphics[width=\linewidth]{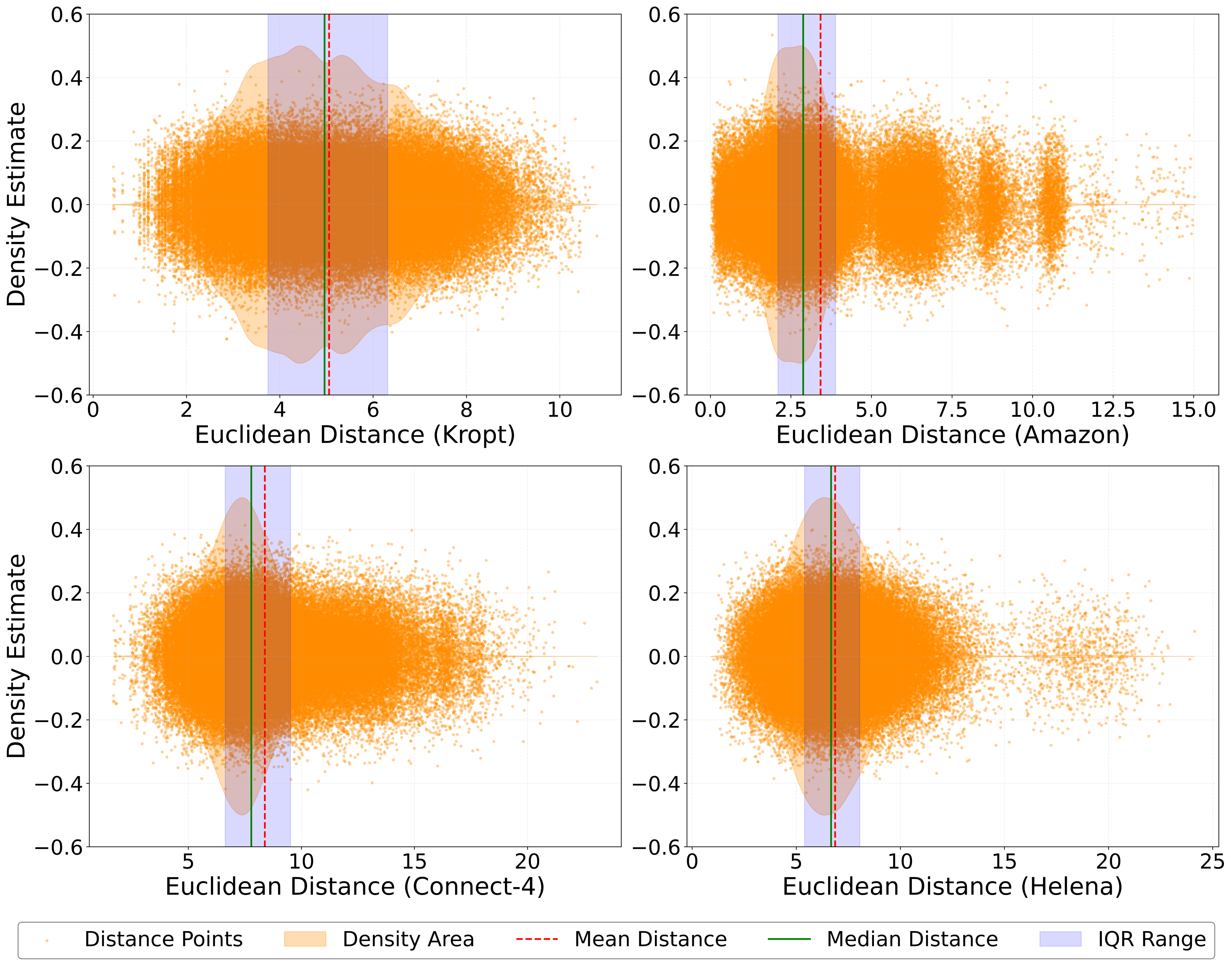}
  \caption{Comparison of sample distance distributions across classification datasets. }
  \label{fig:comparison_cls_distance}
\end{figure}
\vspace{-5pt}
\begin{figure}[htbp]
  \centering
  \includegraphics[width=\linewidth]{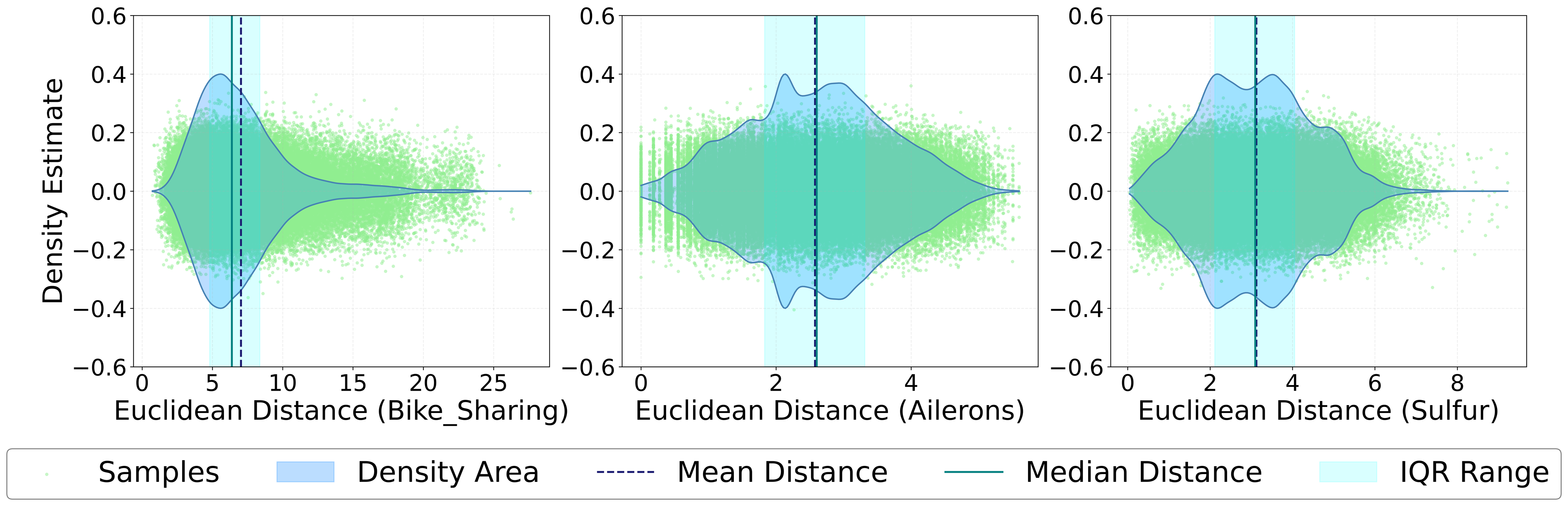}
  \caption{Comparison of sample distance distributions across regression datasets. }
  \label{fig:comparison_reg_distance}
\end{figure}

\vspace{-5pt}

\subsection{The Geometry of Diversity}
\label{geo}
We further analyze how data geometric properties modulate the effectiveness of diversity-aware retrieval, using eight representative datasets spanning 6 to 216 features. Detailed statistics for all eight datasets are provided in Appendix~\ref{appendix:rep} and their performance with TabDivER is summarized in Table~\ref{tab:representative} (TabPFN-v2, $k$=2048). Figures~\ref{fig:comparison_cls_distance} and ~\ref{fig:comparison_reg_distance} show the pairwise distance distributions in the original feature space for a subset of these datasets.

\begin{table}[htbp]
\centering
\caption{Performance on representative datasets.}
\small
\begin{tabular}{@{}l|c|c|c@{}}
\toprule
Dataset & Metric & Baseline & +TabDivER \\
\midrule
Kropt  & ACC & 0.844 $\pm$ 0.006 & \textbf{0.921 $\pm$ 0.004} \\
Amazon  & ACC & 0.927 $\pm$ 0.002 & \textbf{0.934 $\pm$ 0.003} \\
Connect-4  & ACC & 0.810 $\pm$ 0.002 & \textbf{0.838 $\pm$ 0.005} \\
Helena  & ACC & 0.764 $\pm$ 0.003 & \textbf{0.866 $\pm$ 0.003} \\
Bike\_Sharing  & R$^2$ & 0.942 $\pm$ 0.003 & \textbf{0.947 $\pm$ 0.004} \\
Ailerons  & R$^2$ & 0.847 $\pm$ 0.004 & \textbf{0.872 $\pm$ 0.006} \\
Sulfur  & R$^2$ & 0.779 $\pm$ 0.002 & \textbf{0.933 $\pm$ 0.003} \\
Sylva  & ACC & 0.977 $\pm$ 0.001 & \textbf{0.979 $\pm$ 0.002} \\
\bottomrule
\end{tabular}
\label{tab:representative}
\end{table}

Figures~\ref{fig:comparison_cls_distance} and~\ref{fig:comparison_reg_distance} show pairwise distance distributions of the unified feature spaces, revealing two distinct regimes where diversity helps. Datasets with compact distances (e.g., Kropt, Amazon) have narrow interquartile ranges; similarity-only retrieval selects neighbors that are numerically close and thus informationally redundant. Diversity forces the algorithm beyond this dense core, bringing in samples from slightly different regions that provide complementary discriminative evidence, explaining the substantial gains on Kropt (+7.7\%). Datasets with sparse, widely dispersed distances (e.g., Helena, Sulfur) operate differently. Nearest neighbors may already be far from the query, making local similarity unreliable---not due to redundancy but insufficient relevance. Diversity helps by providing global coverage across a multi-modal distribution, correcting decision biases when the model sees only one mode. This accounts for the large gains on Helena (+10.2\%) and Sulfur (+15.4\%).

For regression, similar patterns arise through distinct mechanisms. Unimodal long-tail distributions (e.g., Bike\_Sharing) benefit from diversity via better long-range interpolation, extending the numerical range seen by the model and correcting extrapolation biases. Multi-modal distributions with dense local peaks (e.g., Ailerons) benefit from de-redundancy across clusters, as similarity-only retrieval repeatedly samples the same peak, over-representing one mode. The most pronounced gains occur on sharply multi-modal datasets (e.g., Sulfur), where similarity-only retrieval collapses entirely into a single mode, and diversity acts as a structural integrator forcing cross-cluster sampling to fuse information.

\section{Conclusion and Future Work}
\label{conclusion}
We introduced TabDivER, a plug-and-play diversity-enhanced retrieval framework designed to address the context selection bottleneck in tabular foundation models. By integrating adaptive feature compilation, efficient indexing, and a tabular-specific diversity-adaptive component, TabDivER enables the retrieval of higher-quality contextual samples. Extensive experiments across three benchmarks validate that the framework delivers consistent improvements for state-of-the-art TFMs and exhibits robustness under temporal distribution shift. Furthermore, our analysis of the optimal diversity interval provides insights into balancing relevance and diversity in tabular ICL. Several directions remain for future work. While TabDivER currently prioritizes a geometry-preserving raw feature space to avoid representation collapse, a promising direction is to investigate methods for adaptively integrating high-level metadata. We also plan to extend TabDivER to handle massive-scale, high-velocity streaming tabular data.


\bibliographystyle{ACM-Reference-Format}
\bibliography{ref}

\appendix

\appendix

\section{APPENDIX}

\subsection{Notations}
The notations used in the paper are listed in Table~\ref{tab:summary_notations}.
\begin{table}[!ht]
\caption{\textbf{Summary of notations.}}
\label{tab:summary_notations}
\centering
\footnotesize
\renewcommand{\arraystretch}{1.0} 
\renewcommand{\tabularxcolumn}[1]{m{#1}} 
\begin{tabularx}{\columnwidth}{ 
    >{\centering\arraybackslash}m{0.18\columnwidth} 
    >{\centering\arraybackslash\small}X 
}
\toprule
\multicolumn{1}{c}{\textbf{Symbols}} & \multicolumn{1}{c}{\textbf{Meaning}} \\
\midrule
$D$ & The tabular dataset $D = \{(x_i, y_i)\}_{i=1}^N$ with $N$ samples. \\
$x_{i, k}$ & The $k$-th feature value of the $i$-th sample. \\
$d, d'$ & Original and unified feature dimensionality after filtering. \\
$\tilde{x}_i$ & Unified numeric vector representation in the raw feature space. \\
$\mu_k, \sigma_k$ & Mean and standard deviation of the $k$-th feature. \\
$\phi_k(\cdot)$ & Deterministic ordinal mapping function for categorical features. \\
$\kappa_{\mathrm{clip}}$ & Multiplier for feature-wise $\mu \pm \kappa_{\mathrm{clip}}\sigma$ clipping. \\
$\hat{x}_i$ & $L^2$-normalized feature vector on the unit hypersphere. \\
$q$ & The query instance (test sample) for context retrieval. \\
$\mathcal{I}$ & The Faiss-based exact flat index (\texttt{IndexFlatIP}). \\
$\mathcal{C}_{\mathrm{pre}}$ & Initial set of $M$ candidate samples retrieved by similarity. \\
$S$ & Final set of $K$ samples selected for the in-context window. \\
$sim(u, v)$ & Cosine similarity function in the unified feature space. \\
$N_k(x)$ & $k$-nearest neighbors of sample $x$ in original feature space. \\
$\lambda$ & Relevance--diversity trade-off coefficient (larger values favor relevance). \\
$R(S, q)$ & Average relevance score between the selected set $S$ and query $q$. \\
$D(S)$ & Internal similarity (negative diversity) of samples within $S$. \\
$Penal_L$ & Local diversity penalty based on neighborhood structure. \\
$Concent(q)$ & Similarity distribution concentration of candidates for query $q$. \\
$Complex(q)$ & Local structural complexity (dispersion of top candidates). \\
$\lambda_{base}$ & Baseline relevance weight before query-specific adaptation. \\
\bottomrule
\end{tabularx}
\end{table}

\subsection{Analysis of Representation Spaces}
\label{append:raw space}
To examine the geometric properties relevant to diversity-aware retrieval, we perform a diagnostic comparison on the Kropt dataset between the Unified Raw Feature Space and the TabICL Encoder-induced Embedding Space.

\begin{figure}[htbp]
  \centering
  \includegraphics[width=\linewidth]{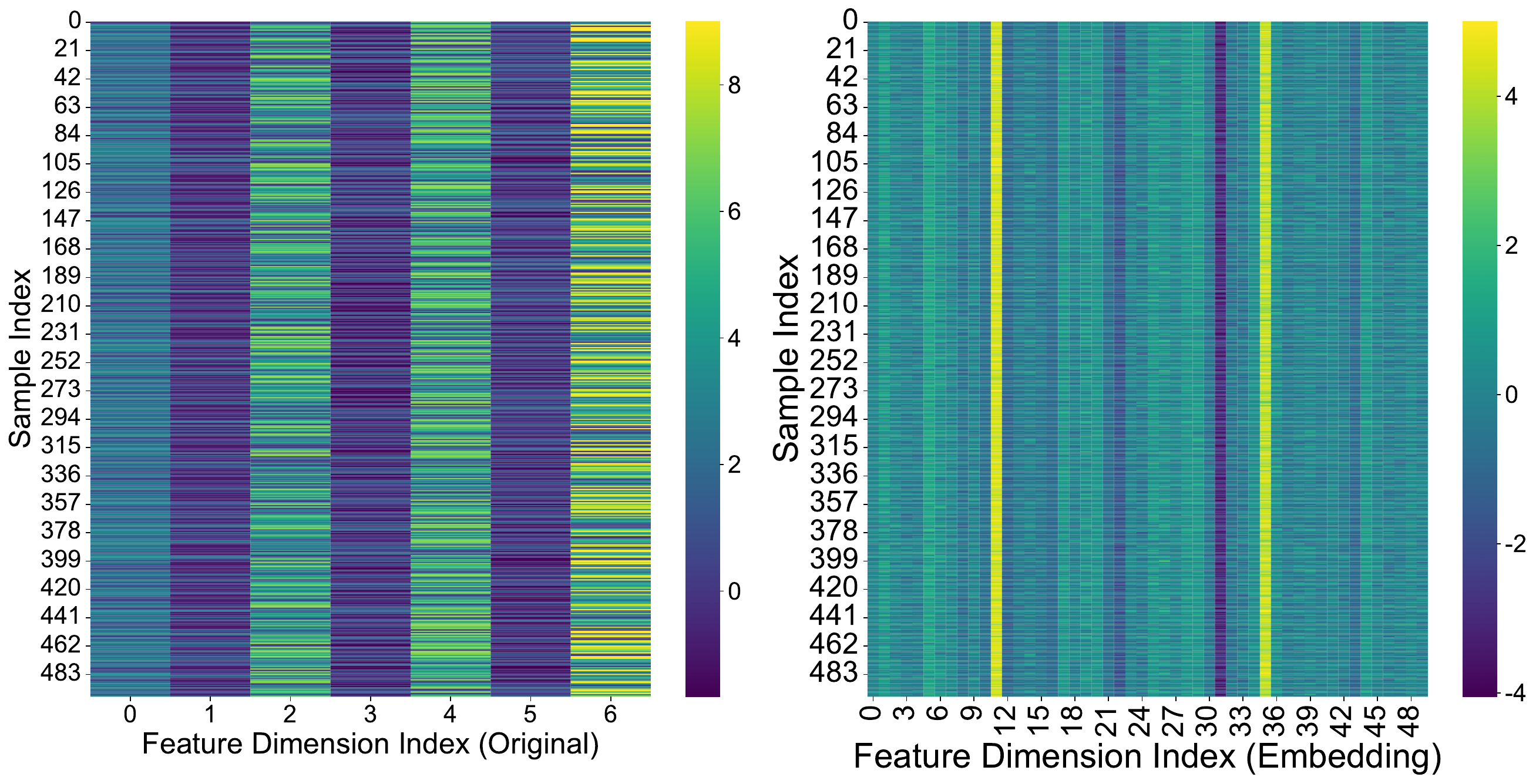}
  \caption{Heatmaps comparison: original space vs embedding space. }
  \label{fig:heatmaps}
\end{figure}

\begin{figure}[htbp]
  \centering
  \includegraphics[width=\linewidth]{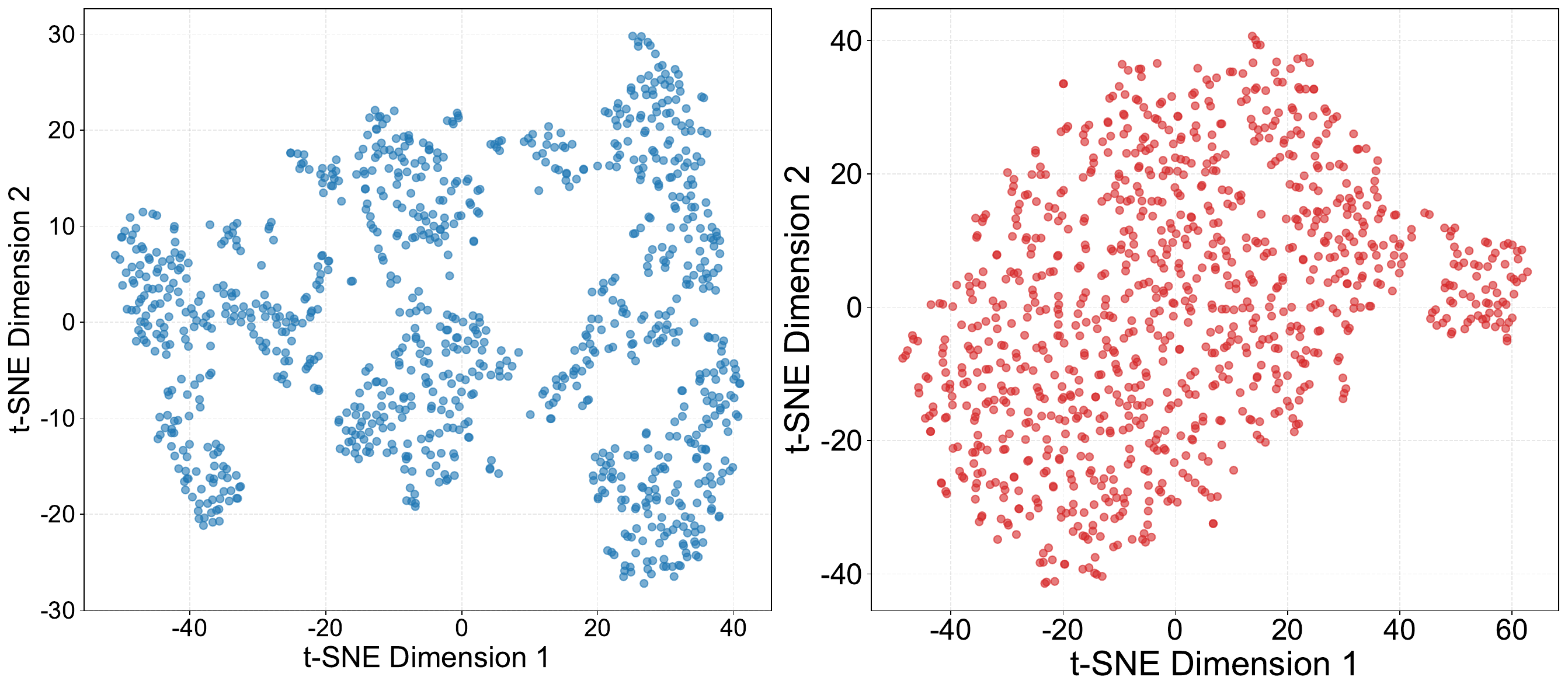}
  \caption{T-SNE comparison: original space vs embedding space. }
  \label{fig:t-sne}
\end{figure}

\noindent\textbf{Structural Sparsity and Interpretability} 
As illustrated in the heatmaps (Figure~\ref{fig:heatmaps}), the raw feature space exhibits more visible feature-wise structure, whereas the 512-dimensional encoder space has more diffuse activations. Table~\ref{tab:comparison_of_statistical} reports the corresponding statistics. These observations motivate using a shared model-agnostic raw feature space, but do not establish that an encoder space is universally unsuitable for retrieval.

\noindent\textbf{Manifold Collapse and Dimensionality} 
PCA results show that $95\%$ of the variance is captured by $6$ components in the raw space and $38$ in the encoder space (Table~\ref{tab:quantitative_comparison}). Given the different ambient dimensions, these counts alone do not prove dimensional collapse. The t-SNE visualization (Figure~\ref{fig:t-sne}) suggests different local grouping patterns; the measured local spread provides complementary dataset-specific evidence about neighborhood separation.
\begin{table}[t]
\small
\centering
\caption{Comparison of statistical indicators.}
\label{tab:comparison_of_statistical}
\renewcommand{\tabularxcolumn}[1]{m{#1}}
\begin{tabularx}{\columnwidth}{>{\centering\arraybackslash}Xcc}
\toprule
Statistical Indicator & Original Space & Embedding Space  \\
\midrule
Overall Sparsity & 0.055 & 0 \\
Mean of Non-zero Values & 2.200 & -0.005 \\
Standard Deviation of Non-zero Values & 2.699 & 1.024 \\
\bottomrule
\end{tabularx}
\end{table}

\begin{table}[t]
\small
\centering
\caption{Quantitative comparison of geometric properties.}
\label{tab:quantitative_comparison}
\renewcommand{\arraystretch}{1.3}
\begin{tabularx}{\columnwidth}{
    >{\centering\arraybackslash\hsize=1.6\hsize}X
    >{\centering\arraybackslash\hsize=0.8\hsize}X
    >{\centering\arraybackslash\hsize=0.8\hsize}X 
    >{\centering\arraybackslash\hsize=0.8\hsize}X
}
\toprule
Metric & Raw Space & Encoder Space & Ratio (Raw/Enc.) \\
\midrule

Concentration Ratio & 3.0 & 9.8 & 0.30x \\
Local Spread Ratio & 2.51 & 1.57 & 1.60x \\

PCA Dimensions (95\%) & 6 & 38 & 0.16x \\
\bottomrule
\end{tabularx}
\end{table}

\noindent\textbf{Concentration of Measure} 
The encoder space has a higher measured concentration ratio than the raw space (approximately $9.8$ vs. $3.0$), while the raw space has a higher local spread ratio ($2.51$ vs. $1.57$). These dataset-specific observations suggest more distinguishable local neighborhoods in the raw feature space, without implying that encoder-space diversity penalties are universally ineffective.

\subsection{Algorithm Description}
The T-MMR algorithm consists of four stages: (1) \textit{Low-Quality Filtering} removes candidates below the query-similarity threshold; (2) \textit{Density Lookup} accesses the precomputed local-density scores; (3) \textit{Adaptive Parameter Tuning} uses the candidate distribution to adjust the query-specific relevance--diversity weight; and (4) \textit{Greedy Diversity Selection} initializes the context with the most relevant example and iteratively adds candidates based on relevance and redundancy. The algorithm returns the selected context for tabular in-context learning.

\begin{algorithm}[htbp]
    \caption{Tabular Diversity-Adaptive Component}
    \label{alg:diversity_component}
    \begin{algorithmic}[1]
        \Require Query sample \( q \), candidate pool \( \mathcal{C}_{\text{pre}} \) from Faiss retrieval, context size \( K \), filtering threshold \( \tau \), base weight \( \lambda_{\text{base}} \), neighborhood size \( k \)
        \Ensure Diversity-enhanced context set \( S \)

        \vspace{0.3em}

        \State \textbf{Stage 1: Low-Quality Filtering}
        \State \(\mathcal{C} \leftarrow \{ c \in \mathcal{C}_{\text{pre}} \mid \operatorname{sim}(q, c) \geq \tau \}\)
        \If{\( |\mathcal{C}| < K \)}
            \State \(\mathcal{C} \leftarrow \mathcal{C}_{\text{pre}}[:K]\) \Comment{Fallback to top-\(K\)}
        \EndIf

        \vspace{0.3em}

        \State \textbf{Stage 2: Density Lookup}
        \For{each \( c \in \mathcal{C} \)}
            \State \(\text{kNN\_density}(c) \leftarrow \operatorname{LookupDensity}(c)\) \Comment{Precomputed offline}
        \EndFor

        \vspace{0.3em}

        \State \textbf{Stage 3: Adaptive Parameter Tuning}
        \State Compute concentration: \(\operatorname{Concent}(q) \leftarrow \mu_{\text{sim}} / \sigma_{\text{sim}}\) over top-\(L\) candidates
        \State Compute complexity: \(\operatorname{Complex}(q) \leftarrow \frac{1}{L} \sum_{i=1}^{L} \| \mathbf{c}_i - \bar{\mathbf{c}} \|_2\)
        \State Apply \(\tanh\) and sigmoid normalization to obtain \(\widehat{\operatorname{Concent}}(q)\) and \(\widehat{\operatorname{Complex}}(q)\)
        \State \(\lambda_{\text{adapt}} \leftarrow \lambda_{\text{base}} \cdot [1 - 0.6 \cdot (1 - \widehat{\operatorname{Concent}}(q)) + 0.8 \cdot \widehat{\operatorname{Complex}}(q)]\)

        \vspace{0.3em}

        \State \textbf{Stage 4: Greedy Diversity Selection}
        \State \(S \leftarrow \emptyset\), \(\lambda \leftarrow \lambda_{\text{adapt}}\)

        \State \textbf{Initialization:} Select the most relevant sample
        \State \(c_1 \leftarrow \arg\max_{c \in \mathcal{C}} \operatorname{sim}(q, c)\)
        \State \(S \leftarrow \{c_1\}\), \(\mathcal{C} \leftarrow \mathcal{C} \setminus \{c_1\}\)
        \State \(\operatorname{penal\_set}(c) \leftarrow \operatorname{sim}(c,c_1),\; \forall c \in \mathcal{C}\) \Comment{Initialize running maxima}

        \State \textbf{Iterative Expansion:}
        \For{\(t = 2\) \textbf{to} \(K\)}
            \State \textbf{Marginal Gain:} For each \(c \in \mathcal{C}\), compute:
            \State \quad \(\operatorname{penal}(c) \leftarrow \max(\operatorname{penal\_set}(c), \operatorname{kNN\_density}(c))\)
            \State \quad \(\operatorname{gain}(c) \leftarrow \lambda \cdot \operatorname{sim}(q, c) - (1 - \lambda) \cdot \operatorname{penal}(c)\)
            \State \(c^* \leftarrow \arg\max_{c \in \mathcal{C}} \operatorname{gain}(c)\)
            \State \(S \leftarrow S \cup \{c^*\}\), \(\mathcal{C} \leftarrow \mathcal{C} \setminus \{c^*\}\)
            \State \textbf{Incremental Update:} For each remaining \(c \in \mathcal{C}\):
            \State \quad \(\operatorname{penal\_set}(c) \leftarrow \max(\operatorname{penal\_set}(c), \operatorname{sim}(c, c^*))\) \Comment{\(\mathcal{O}(1)\) per candidate}
        \EndFor

        \vspace{0.3em}

        \State \textbf{return} Diversity-enhanced context set \(S\) of size \(K\)
    \end{algorithmic}
\end{algorithm}

\subsection{Analysis of the Expansible Condition}\label{append:analysis}
For diverse context selection, it is desirable that successive neighbors provide geometric variation. Let $x_{(k)}$ denote the $k$-th nearest neighbor of $x$, and let $R_k(x) = d(x,x_{(k)})$ be its distance. A useful geometric desideratum is an appreciable average neighbor gap (an ``expansible neighborhood''):
\begin{equation}
\Delta_k = \mathbb{E}_x [R_{k+1}(x) - R_k(x)] > \delta, \quad \delta > 0 ,
\end{equation}
where $\Delta_k$ represents the expected marginal geometric gap. 

\paragraph{Interpretation.}
A small measured neighbor gap indicates that successive neighbors occur at similar distances, which can limit the geometric distinction available to the reranker. However, lower effective dimensionality does not by itself imply that the gap vanishes: the distances also depend on local density, sampling, and geometry. We therefore treat local neighborhood expansibility as a diagnostic motivation rather than a mathematical guarantee.

\subsection{Complexity Analysis of Algorithm}\label{appendix:complexity}

We analyze the time complexity of the diversity-adaptive greedy selection algorithm. Let $M$ be the candidate pool size and $K$ the target context size.

\noindent\textbf{Precomputation.} Local neighborhood density, $\text{kNN\_density}(c)=\frac{1}{k}\sum_{n\in N_k(c)}\operatorname{sim}(n,c)$, is precomputed using exact Faiss self-queries and reused for reranking. For a flat index of all $N$ training vectors, evaluating every indexed vector against all $N$ vectors requires $\mathcal{O}(N^2d')$ similarity arithmetic offline. The per-query computation described below excludes this one-time step.

\noindent\textbf{Nominal complexity.} At iteration $t$ ($1 \leq t \leq K$), the algorithm evaluates all remaining $M - t + 1$ candidates. For each candidate, computing $\max_{s \in S} \text{sim}(c, s)$ requires comparing against the current selected set of size $t-1$. Thus the cost at iteration $t$ is $(M - t + 1) \cdot (t - 1)$. Summing over all iterations:
\[
\sum_{t=1}^{K} (M - t + 1)(t - 1) = \mathcal{O}(M K^2).
\]

\noindent\textbf{Optimization via incremental updates.} The $\mathcal{O}(M K^2)$ complexity can be reduced to $\mathcal{O}(M K)$ by maintaining for each remaining candidate $c$ the current maximum similarity to any selected sample. When a new sample $c_t$ is added to $S$, each remaining candidate updates its penalty as:
\[
\text{penal\_set}(c) \leftarrow \max(\text{penal\_set}(c), \text{sim}(c, c_t)),
\]
requiring $\mathcal{O}(1)$ for the maximum update once the similarity is available. This eliminates rescanning the selected set $S$ and yields $\mathcal{O}(MK)$ score updates.

\noindent\textbf{Online reranking.}
By incrementally maintaining candidate redundancy scores,
the greedy reranker requires $\mathcal{O}(MK)$ score updates
per query, in addition to the $\mathcal{O}(Nd')$ exact
candidate search (Section~\ref{sec:index}).
Reranking is performed using batched GPU operations,
while density estimation is precomputed offline.

\subsection{Dataset Statistics for Representative Subset}
\label{appendix:rep}
\begin{table}[htbp]
\centering
\caption{Dataset-level statistics for the eight representative datasets .}
\small
\begin{tabular}{@{}l|l|c|c|c@{}}
\toprule
No. & Dataset & \#Features & \#Samples & Task \\
\midrule
1 & Kropt & 6 & 28,056 & Classification \\
2 & Amazon\_employee\_access & 9 & 32,769 & Classification \\
3 & Connect-4 & 42 & 67,557 & Classification \\
4 & Helena & 27 & 65,196 & Classification \\
5 & Bike\_Sharing\_Demand & 11 & 17,379 & Regression \\
6 & Ailerons & 33 & 13,750 & Regression \\
7 & Sulfur & 6 & 10,081 & Regression \\
8 & Sylva & 216 & 14,394 & Classification \\
\bottomrule
\end{tabular}
\label{tab:dataset_stats_appendix}
\end{table}

\subsection{Implementation Details}
\label{appendix:implementation}
\noindent\textbf{Dataset splitting and backbone compatibility.}
Each dataset uses an $80\%$ context/retrieval pool and a held-out $20\%$ test set, with test labels used only for evaluation. For data exceeding backbone class or sample limits, we retain the 10 largest classes where required and use class-balanced stratified subsampling to at most 10{,}000 training examples. These are backbone-compatibility adjustments rather than dataset-selection criteria; all compared retrieval strategies use the same backbone-specific constraints.

\noindent\textbf{Integration and parameter selection.}
TabDivER replaces the native retriever for TabDPT and LimiX and externally selects the context for TabPFN-v2 and TabICL without updating backbone parameters. The adaptive weighting rule is shared across datasets, with no dataset-specific selection based on validation or test results. The $\lambda$ grids are used only for sensitivity analysis. The candidate pool size is $M=\min\{N,15K\}$, where $N$ denotes the number of available retrieval samples.

\noindent\textbf{Implementation and evaluation.}
The implementation uses Python, PyTorch, Faiss, NumPy, and Pandas, with exact \texttt{IndexFlatIP} candidate search and batched GPU reranking on an NVIDIA RTX 5090. The eight representative datasets in Appendix~\ref{appendix:rep} were evaluated over three runs for focused analysis. Main benchmark results are computed per task and aggregated within each suite. Throughput is measured under batched GPU execution, allowing concurrent processing of multiple queries; it is not the reciprocal of single-query latency.

\end{document}